\pdfoutput=1
\documentclass[11pt]{article}

\usepackage{acl}

\usepackage{times}
\usepackage{latexsym}

\usepackage[T1]{fontenc}
\usepackage[utf8]{inputenc}

\usepackage{microtype}

\usepackage{graphicx}
\usepackage{amsmath,amsfonts,bm}
\usepackage{multirow}
\usepackage{booktabs}
\usepackage{CJKutf8}
\usepackage{subfigure}
\usepackage{paralist}
\newcommand{\tabref}[1]{\tablename~\ref{#1}}
\newcommand{\figref}[1]{\figurename~\ref{#1}}

\title{Identifying Chinese Opinion Expressions with Extremely-Noisy Crowdsourcing Annotations}

\author{
Xin Zhang$^1$, Guangwei Xu, Yueheng Sun$^2$, Meishan Zhang$^{3}$\thanks{~~Corresponding author.} , Xiaobin Wang, Min Zhang$^3$\\
$^1$School of New Media and Communication, Tianjin University, China\\
$^2$College of Intelligence and Computing, Tianjin University, China\\
$^3$Institute of Computing and Intelligence, Harbin Institute of Technology (Shenzhen), China\\
\texttt{\{hsinz,yhs\}@tju.edu.cn, \{ahxgwOnePiece, mason.zms\}@gmail.com}\\
\texttt{czwangxiaobin@foxmail.com, zhangmin2021@hit.edu.cn}
}

\begin{document}
\begin{CJK}{UTF8}{gbsn}

\maketitle
\begin{abstract}
Recent works of opinion expression identification (OEI) rely heavily on the quality and scale of the manually-constructed training corpus, which could be extremely difficult to satisfy. % especially for the languages such as Chinese.
Crowdsourcing is one practical solution for this problem, aiming to create a large-scale but quality-unguaranteed corpus.
In this work, we investigate Chinese OEI with extremely-noisy crowdsourcing annotations, constructing a dataset at a very low cost.
Following \newcite{zhang-etal-2021-crowdsourcing}, we train the annotator-adapter model by regarding all annotations as gold-standard in terms of crowd annotators, and test the model by using a synthetic expert, which is a mixture of all annotators.
As this annotator-mixture for testing is never modeled explicitly in the training phase,
we propose to generate synthetic training samples by a pertinent mixup strategy to make the training and testing highly consistent.
The simulation experiments on our constructed dataset show that crowdsourcing is highly promising for OEI, and our proposed annotator-mixup can further enhance the crowdsourcing modeling.
\end{abstract}

\section{Introduction}
\begin{figure}
\centering
\includegraphics[scale=0.753]{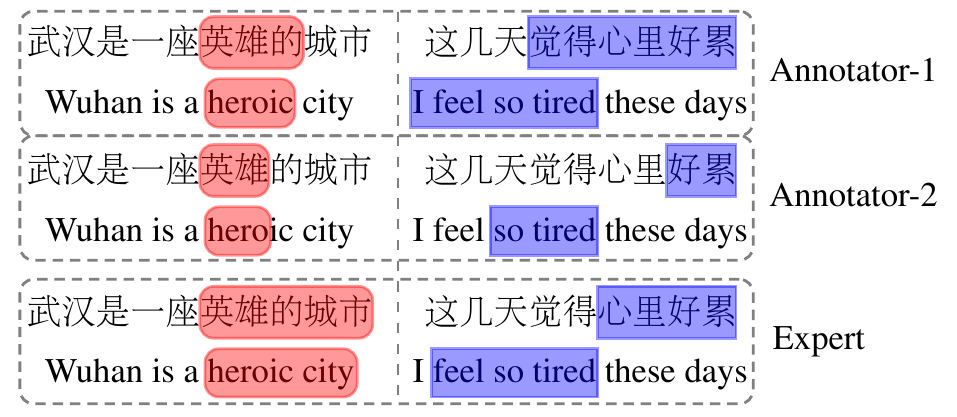}
\caption{
Two examples of opinion expression identification with crowdsourcing and expert annotations in our constructed dataset. The left and right sentences are of positive and negative polarities, respectively.
}
\label{figure:example}
\end{figure}

% 1
Opinion mining is a fundamental topic in the natural language processing (NLP) community, which has received great attention for decades \cite{liu2012survey}.
Opinion expression identification (OEI) is a standard task of opinion mining, which aims to recognize the text spans that express particular opinions \cite{breck2007identifying}.
\figref{figure:example} shows two examples.
This task has been generally solved by supervised learning \cite{irsoy-cardie-2014-opinion} with the well-established corpus annotated by experts.
Almost all previous studies are based on English datasets such as MPQA \cite{wiebe2005annotating}.

% 2
By carefully examining this task, we can find that the corpus annotation of opinion expressions is by no means an easy process.
It is highly ambiguous across different persons.
As shown in \figref{figure:example}, it is very controversial to define the boundaries of opinion expressions \cite{wiebe2005annotating}.
Actually, this problem is extremely serious for languages such as Chinese, which is based on characters even with no explicit and clearly-defined word boundaries.
Thus, Chinese-alike languages will inevitably involve more ambiguities.

% 3
In order to obtain a high-quality corpus, we usually need to train the annotators with great efforts, making them acquainted with a specific fine-grained guideline drafted by experts, and then start the data annotation strictly.
Finally, it is better with a further expert checking on borderline cases where the annotators disagree most to ensure the quality of the annotated corpus.
Apparently, the whole process is quite expensive.
Thus, crowdsourcing with no training (just a brief guideline) and no expert checking is more practical in real considerations \cite{snow-etal-2008-cheap}.
While on the other hand, the difficulty of the Chinese OEI task might lead to very low-quality annotations by crowdsourcing.

% 4
In this work, we present the first study of Chinese OEI by using crowdsourcing.
We manually construct an OEI dataset by crowdsourcing, which is used for training.
Indeed, the dataset is cheap but with a great deal of noises according to our initial observation.
We also collect the small-scale development and test corpus with expert annotations for evaluation.\footnote{
In addition, we provide expert annotations of trainset to train a upper-bound model.
}
Our dataset is constructed over a set of Chinese texts closely related to the COVID-19 topic.
Following, we start our investigation by using a strong BERT-BiLSTM-CRF model,
treating the OEI task as a standard sequence labeling problem following the previous studies \cite{breck2007identifying,irsoy-cardie-2014-opinion,katiyar-cardie-2016-investigating}.
{Our primary goal is to answer whether these extremely-noisy crowdsourcing annotations include potential value for the OEI task.}

% 5
In order to make the best use of our crowdsourcing corpus, we follow \newcite{zhang-etal-2021-crowdsourcing} to treat all crowd annotations as gold-standard in terms of different annotators.
We introduce the annotator-adapter model, which employs the crowdsourcing learning approach of \newcite{zhang-etal-2021-crowdsourcing} in OEI for the first time.
% \zx{这句怎么清晰一下}
It jointly encodes both texts and annotators, then predicts the corresponding crowdsourcing annotations in the BERT-BiLSTM-CRF architecture.
Concretely, we train the annotator-adapter model by each individual annotator and the corresponding annotations, then test the model by using a pseudo expert annotator, which is a linear mixture of crowd annotators.
Considering that this expert is never modeled during the training, we further exploit a simple mixup \cite{zhang2018mixup} strategy to simulate the expert decoding accurately.

% 6
Experimental results show that crowdsourcing is highly competitive, giving an overall F1 score of $53.86$ even with a large-scale of noises, while the F1 score of expert corpus trained model is $57.08$.
We believe that this performance gap is totally acceptable for building OEI application systems.
In addition, our annotator-mixup strategy can further boost the performance of the annotator-adapter model, giving an F1 increase of $54.59-53.86 = 0.73$.
We conduct several analyses to understand the OEI with crowdsourcing and our suggested methods comprehensively.

In summary, we make three majoring contributions as a whole in this work:
\begin{compactitem}
\item We present the initial work of investigating the OEI task with crowdsourcing annotations, showing its capability on Chinese.
\item We construct a Chinese OEI dataset with crowdsourcing annotations, which is not only valuable for Chinese OEI but also instructive for crowdsourcing researching.
\item We introduce the annotator-adapter for crowdsourcing OEI and propose the annotator-mixup strategy, which can effectively improve the crowdsourcing modeling.
\end{compactitem}
All of our codes and dataset will be available at \href{https://github.com/izhx/crowd-OEI}{github.com/izhx/crowd-OEI} for research purpose.

\begin{table}[tb]
\centering 
\resizebox{0.43\textwidth}{!}{
\begin{tabular}{cc}
\toprule
No. & Chinese / English \\
\midrule
\multirow{3}{*}{1} & 澳大利亚籍返京女子不隔离外出跑步 / \\
 & The Australian woman running outside \\
 & without isolation in Beijing \\
2 & 单玉厚 / Yuhou Shan \\
3 & 李文亮医生 / Dr. Li Wenliang \\
4 & 是谁发现了病毒 / Who finds the virus \\
5 & 方方日记 / Fang Fang's Diary \\
6 & 歌诗达赛琳娜号 / Goethe Serena \\
\multirow{2}{*}{7} & 新冠可通过气溶胶传播 / \\
& COVID-19 can transmit via aerosol \\
\bottomrule
\end{tabular}
}
\caption{Seven hot topics we targeted.}
\label{table:topics}
\end{table}

\section{Dataset}\label{dataconstruct}
The outbreak of COVID-19 brings strong demand for building robust Chinese opinion mining systems, which are practically built in a supervised manner.
A large-scale training corpus is the key to the system construction, while almost all existing related datasets are in English \cite{wiebe2005annotating}.
Hence, we manually construct a Chinese OEI dataset by crowdsourcing.
We focus on opinion expressions with {\bf positive} or {\bf negative} polarities only.
The construction consists of four steps: (1) text collection, (2) annotator recruitment, (3) crowd annotation, and (4) expert checking and correction.

\subsection{Text Collection}
We choose the Sina Weibo\footnote{\href{https://weibo.com}{https://weibo.com}}, which is a Chinese social media platform similar to Twitter, as our data source.
To collect the texts strongly related to COVID-19,
we select around 8k posts that are created from January to April 2020 and related to seven hot topics (\tabref{table:topics}).
To make these posts ready for annotating, we use HarvestText\footnote{
\href{https://github.com/blmoistawinde/HarvestText}{https://github.com/blmoistawinde/HarvestText}} to clean them and segment the resulting texts into sentences.
Next, we conduct another cleaning step to remove the duplicates and sentences with relatively poor written styles (e.g., high-proportion of non-Chinese symbols, very short /long length, etc.).

After the above procedure, there are still a large proportion of sentences that involve no sentiment. So we filter out them by a BERT sentiment classifier that trained on an open-access Weibo sentiment classification dataset.\footnote{
\href{https://github.com/SophonPlus/ChineseNlpCorpus/blob/master/datasets/weibo_senti_100k/intro.ipynb}{ChineseNlpCorpus - weibo\_senti\_100k}
}
Only sentences with high confidence of not expressing any sentiment are dropped,\footnote{
Note that there are still a small number of sentences in our final dataset that have no opinion expression inside.
} we can therefore keep the most valuable contents while avoiding unnecessary annotations and thus reduce the overall annotating cost.

% 没有guideline，或者尽量讲的简单。专家没有gudeline，只有crowdsourcing有guide
% 不要写labmates，只写有情感分析经验的人。
\subsection{Annotator Recruitment}
We have five professionals who have engaged in the annotation of sentiment and opinion-related tasks previously and are with rich experience as experts.
They annotate 100 sentences together as examples (i.e., label the positive and negative opinion expressions inside the texts), and establish a simple guideline based on their consensus after several discussions.
The guideline includes the task definition and a description of annotation principle.\footnote{
We share the guideline in the Appendix \ref{appendix:guideline}.
}

Next, we recruit 75 (crowd) students in our university for annotating.
They come from different grades and different majors, such as Chinese, Literature, and Translation.
We offer them the above annotation guideline to understand the task.
We choose the doccano\footnote{
\href{https://github.com/doccano/doccano}{https://github.com/doccano/doccano}
} to build up our annotation platform, and let these annotators be familiar with our task by the expert-annotated examples.
% \footnote{
% Annotating this dataset requires the annotator very familiar with the Chinese language, culture and current events.
% The existing crowdsourcing platforms, such as Amazon Mechanical Turk, might not have this kind of workers available since they are often not open registration to China Mainland.
% Hence we build the platform and recruit crowd workers by ourselves.
% }

\subsection{Crowd Annotation}
When all crowd workers are ready, we start the crowd annotation phase. The prepared texts are split into micro-tasks so that each one consists of 500 sentences.
Then we assign 3 to 5 workers to each micro-task, and their identities are remained hidden from each other.
Each worker will not access a new task unless their current one is finished.

In the annotation of each sentence, workers need to label the positive and negative opinion expressions according to the guideline and their understandings.
The number of positive or negative expressions in one sentence has no limit.
They can also mark a sentence as ``No Opinion'' and skip it if they think there are no opinion expressions inside.

\subsection{Expert Checking and Correction}\label{data:silver}
After all crowd annotations are accomplished, we randomly select a small proportion of sentences and let experts reannotate them, resulting in the {\bf gold-standard} {\it development} and {\it test} corpus.\footnote{
The corresponding crowdsourcing annotations consist of the {\bf crowdsourcing} {\it development} and {\it test} corpus.
}
Specifically, for each sentence, we let 2 experienced experts individually reannotate it with references from the corresponding crowdsourcing annotations.
They will give the final annotation of each sentence if their answers reach an agreement. And if they have divergences, a third expert will help them to modify answers and reach the agreement.
% \footnote{
% We also annotated the {\bf gold-standard} {\it training} corpus.
% }

Then, we let all five experts go through the remaining dataset\footnote{
The remaining part is the {\bf crowdsourcing} {\it training} corpus.
}, selecting the best annotations for each sentence, which can be regarded as the {\bf silver-standard} {\it training} corpus.
In the selection, Each sentence is assigned to 1 expert, and the expert is only allowed to choose one (or several identical) best answer(s) from all the candidate crowdsourcing annotations.
Finally, only for comparisons, we also annotated the {\bf gold-standard} {\it training} corpus, which will not be used in our model training.

\begin{table}[tb]
\centering
\resizebox{0.48\textwidth}{!}{
\renewcommand\arraystretch{0.8}
\setlength{\tabcolsep}{2.4pt}
\begin{tabular}{ccrrrc}
\toprule
\multicolumn{2}{c}{Dataset} & \multicolumn{3}{c}{Number of} & \multirow{3}{1.25cm}{Average Span Length}  \\ \cmidrule(lr){1-2} \cmidrule(lr){3-5}
% \specialrule{0em}{1pt}{1pt}
\multirow{2}{*}{Section} &  \multirow{2}{*}{Quality} & Unique & Positive & Negative & \\
& & Annotation & Expression & Expression & \\
\specialrule{0em}{1pt}{2pt}
\midrule
\multirow{3}{*}{Train}  &  crowd & 32582  & 11640 & 35263 & 5.05 \\
\specialrule{0em}{1pt}{1pt}
                        & silver & 8047  & 4167  & 11411 & 4.71 \\
\specialrule{0em}{1pt}{1pt}
                        & gold & 8047  & 3488  & 10096 & 4.79 \\
\cmidrule(){1-6}
\multirow{2}{*}{Dev}  & crowd & 3427 &   2338  & 3905  & 5.22 \\
\specialrule{0em}{1pt}{1pt}
                        & gold  & 803  &   706  & 1035  & 5.02 \\
\cmidrule(){1-6}
\multirow{2}{*}{Test} & crowd & 6265 &   3573  & 5290  & 4.48 \\
\specialrule{0em}{1pt}{1pt}
                        & gold  & 1517 &   999  & 1373  & 4.30 \\
\bottomrule
\end{tabular}
}
\caption{
Data statistics of our constructed dataset.
For gold and silver corpus, each annotation corresponds to one sentence.
For the crowd corpus, each sentence has 3 to 5 annotations.
So we have a total number of $803 + 1517 + 8047 = 10,367$ unique sentences and $32,582 + 3427 + 6265 = 42,274$ crowd annotations.
}
\label{table:dataset}
\end{table}

% 在dataset部分强调数据还是有用的
\subsection{Dataset Statistics}
In the end, we arrive at $42,274$ crowd annotations by 70 valid annotators,\footnote{
We removed 5 annotators who gave up this work in their first assigned task as a basic quality assurance.
} covering $10,367$ sentences.
A total number of $803 + 1517 = 2320$ sentences, including expert annotations, would be used for development and test evaluations.
\tabref{table:dataset} shows the overall data statistics.
The average number of annotators per sentence is $4.05$, and each annotator labels an average of $827$ sentences in the whole corpus.
The overall Cohen's Kappa value of the crowd annotations is $0.35$.
When ignoring the characters which no annotators think that they are in any expression, the Kappa is only $0.17$.\footnote{
To compute the Kappa value of sequential annotations, we treat each token (not sentence) as an instance, and then aggregate the results of one sentence by averaging.}

The Kappa values are indeed very low, indicating the great and unavoidable ambiguities of the task with natural annotations.\footnote{
The average value of F1 scores that each annotator against the expert is $41.77\%$, which is significantly lower than $60\%+$ of crowdsourcing NER dataset \cite{rodrigues2014sequence}.
}
However, these values do not make much sense since we do not impose any well-designed comprehensive guidelines during annotation.
% According to the statistics, the number of negative opinions is much larger than the positive opinions, which is very interesting and reasonable since COVID-19 brings many negative effects to our daily life.
In fact, a comprehensive guideline for crowd workers is almost impracticable in our task, because they are quite often to disagree with a particular guideline by their own unique and naive understandings.
If we impose such a guideline to them forcibly, the annotation cost would be increased drastically (i.e., at least ten times more expensive according to our preliminary investigation) for their reluctance as well as endless expert guidance.
In the remaining of this work, we will try to verify the real value of these crowdsourcing annotations empirically: Is the collected training corpus really beneficial for our Chinese OEI task?
%When we apply fewer restrictions to the annotating, we can hire workers at a lower price and will afford lower costs.
%In crowdsourcing annotating, we could focus on the knowledge in labels and don't care about the overall annotation quality.
%This is the value of crowdsourcing and makes it possible to construct large-scale corpus in the future.

%\zx{强调了low cost，得加上成本的描述}

\section{Methodology}
The OEI task aims to extract all polarized text spans that express certain opinions in a sentence.
It can be naturally converted into a sequence labeling problem by using the BIO schema, tagging each token by the boundary information of opinion expressions, where ``B-X'' and ``I-X'' (i.e., ``X'' can be either ``POS'' or ``NEG'' denoting the polarity) indicate the start and other positions of a certain expression, and ``O'' denotes a token do not belong to any expression.
In this work we adopt the CRF-based system \cite{breck2007identifying} to the neural setting and enhance it with BiLSTM encoder as well as pre-trained BERT representation.

\subsection{BERT-BiLSTM-CRF Baseline}
\label{baseline}
Given a sentence $\bm{x} = x_1\cdots x_n$ (where $n$ denotes the sentence length),
we first convert it into contextual representations $\bm{r}_1 \cdots \bm{r}_n$
by the pre-trained BERT with adapter tuning \cite{houlsby2019parameter}:
\begin{equation}
\label{bert_repr}
\begin{split}
\bm{r}_1 \cdots \bm{r}_n &= \text{ADBERT}(x_1\cdots x_n). \\
\end{split}
\end{equation}

\begin{figure}
\centering
\subfigure{\includegraphics[scale=0.781]{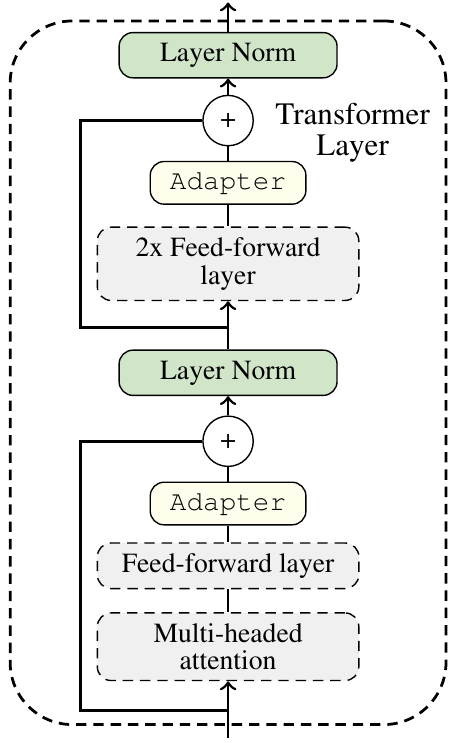}}
\subfigure{\includegraphics[scale=0.781]{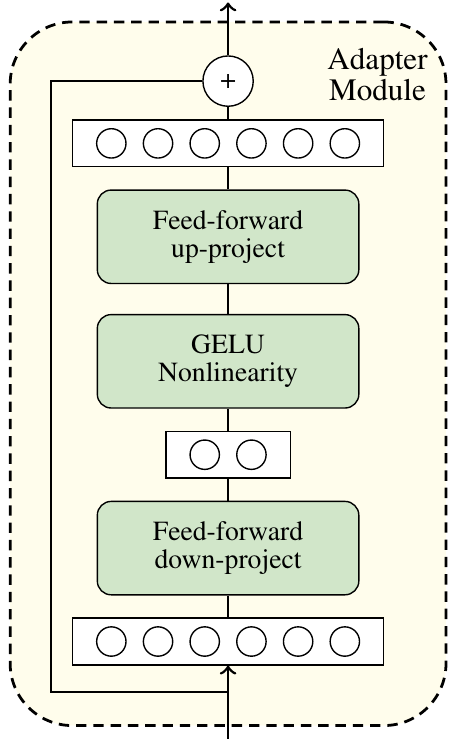}}
\caption{
The Adapter (right) and Transformer integrated with Adapter inside (left).
% {\bf Left}: We add the adapter module after two LayerNorms to each Transformer layer.
% {\bf Right}: The adapter contains:
% (1) a down-projection to filter bottleneck features,
% (2) a non-linearity and an up-projection for reconstructing features to the original size,
% and (3) a skip-connection to improve the convergence.
During the adapter tuning, green layers are trainable, including the adapters,
the LayerNorm, and other task-specific modules.
}
\label{fig:transformer}
\end{figure}

Unlike the standard BERT exploration, % which lets the input go through BERT directly,
ADBERT introduces two extra adapter modules inside each transformer layer, as shown in \figref{fig:transformer} for the details.
With this modification, we do not need fine-tuning all BERT parameters,
and instead, learning the parameters of adapters is enough for obtaining a strong performance.
Thus ADBERT is more parameter efficient.
The standard adapter layer can be formalized as:
\begin{equation} \label{adapter}
\begin{split}
& \text{down-proj:~~~~} \bm{h}_{\text{mid}} =
    \mathrm{GELU}(\bm{W}_{\text{down}}\bm{h}_{\text{in}} +
        \bm{b}_{\text{down}}) ,\\
& \text{up-proj:~~~~~~~~~} \bm{h}_{\text{out}} =
    \bm{W}_{\text{up}}\bm{h}_{\text{mid}} + \bm{b}_{\text{up}} +
    \bm{h}_{\text{in}},
\end{split}
\end{equation}
where $\bm{W}_{\text{down}}$, $\bm{W}_{\text{up}}$, $\bm{b}_{\text{down}}$ and
$\bm{b}_{\text{up}}$ are model parameters, which are much smaller than the
parameters of transformer in scale, and the dimension size of
$\bm{h}_{\text{mid}}$ is also smaller than that of the corresponding transformer
dimension.\footnote{
The dimension sizes of $\bm{h}_{\text{in}}$ and $\bm{h}_{\text{out}}$ are consistent with the corresponding transformer hidden states.
}

The rest part of the baseline is a standard BiLSTM-CRF model, which is a stack of BiLSTM, MLP and CRF layers,
and then we can obtain sequence-level scores for each candidate output $\bm{y}$:
\begin{equation}
\begin{split}
\mathrm{score}(\bm{y}) &= \text{BiLSTM-CRF}([\bm{r}_1\cdots \bm{r}_n]), \\
p(\bm{y}) &= \frac{\exp\big(\mathrm{score}(\bm{y})\big)}{\sum_{\tilde{Y}}
    \exp\big(\mathrm{score}(\tilde{\bm{y}})\big)} , \\
\end{split}
\end{equation}
where $p(\bm{y})$ is the probability of the given ground-truth, and $\widetilde{Y}$ is all possible outputs for score normalization.
The model parameters are updated by the sentence-level cross-entropy loss $\mathcal{L} = -\log p(\bm{y}^*)$ when $\bm{y}^*$ is regarded as gold-standard.

\paragraph{Crowdsourcing training.}
In the crowdsourcing setting, we only have annotations from multiple non-expert annotators, thus no gold-standard label is available for our training.
To handle the situation, we introduce two straightforward and widely-used methods.
First, we treat all annotations uniformly as training instances, despite that they may offer noises for our training objective, which is denoted by \texttt{All} for short.
Second, we exploit majority voting\footnote{
The voting is conducted at the token-level and then merge continuous tokens if they belong to a same-type expression.
} to obtain an aggregated answer of each sentence for model training, denoted as \texttt{MV}.
%Both the two have been widely exploited as baseline in the crowdsource investigation.

\subsection{Annotator Adapter}
\label{sec:annotator-adapter}
\begin{figure}
\centering
\includegraphics[scale=0.93]{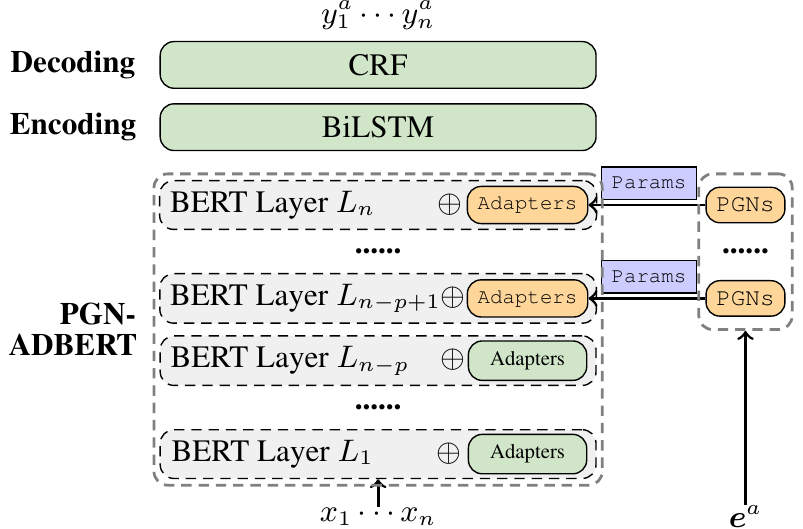}
\caption{The annotator-adapter model. Given a joint input of the text
$x_1\cdots x_n$ and the annotator ID $a$, we first convert $a$ to its embedding
$\bm{e}^a$. Then, PGN use $\bm{e}^a$ generate annotator-specific parameters for
the adapters in top $p$ BERT layers (i.e., from $L_n$ to $L_{n - p + 1}$) to
compute annotator-aware input representations. Finally, the BiLSTM encode
the representations to high-level features and the CRF decoder predict the
labels $y_1^a \cdots y_n^a$ that $a$ gives to $x_1\cdots x_n$.}
\label{fig:model}
\end{figure}

In most previous crowdsourcing studies, there is a common agreement that crowd annotations are noisy, which should be rectified during training \cite{rodrigues2014gaussian,nguyen-etal-2017-aggregating,simpson-gurevych-2019-bayesian}.
\newcite{zhang-etal-2021-crowdsourcing} propose to regard all crowdsourcing annotations
as gold-standard, and introduce a representation learning model to jointly
encode the sentence and the annotator and extract annotator-aware features,
% achieving state-of-the-art performance on the benchmark crowdsourcing NER dataset.
which models the unique understandings of annotators (this setting is indeed very consistent with our corpus).
Since our constructed dataset has no gold-standard training labels\footnote{
We have added the gold-standard annotations in the revision of this work, but we keep this data setting.
}, we adopt their unsupervised representation learning approach, %\cite{zhang-etal-2021-crowdsourcing} to our task
which is named \texttt{annotator-adapter}.
It applies the Parameter Generator Network (PGN) \cite{platanios-etal-2018-contextual,jia-etal-2019-cross,ustun-etal-2020-udapter} to generate annotator-specific adapter parameters for the ADBERT, as shown in Figure \ref{fig:model}.

Given an input sentence-annotator pair $(\bm{x} = x_1, \dots, x_n, a)$, we exploit an embedding layer to convert the annotator ID $a$ into its vectorial form $\bm{e}^a$, and then PGN is used to generate the model parameters of several high-level adapter layers inside BERT conditioned by $\bm{e}^a$.
Concretely, we apply PGN to the last $p$ layers of BERT, where $p$ is one
hyper-parameter of our model.
We refer to PGN-ADBERT for the updated input representation.

Formally, for an adapter defined by Equation \ref{adapter},
all its parameters are dynamically generated by:
\begin{equation}
\begin{split}
    & \bm{W}_{\text{down}} = \bm{T}_{\bm{W}_{\text{down}}} \times \bm{e}^a,
    ~~~~\bm{b}_{\text{down}} = \bm{T}_{\bm{b}_{\text{down}}} \times \bm{e}^a, \\
    & \bm{W}_{\text{up}} = \bm{T}_{\bm{W}_{\text{up}}} \times \bm{e}^a,
    ~~~~\bm{b}_{\text{up}} = \bm{T}_{\bm{b}_{\text{up}}} \times \bm{e}^a, \\
\end{split}
\end{equation}
where $\bm{T}_{\bm{W}_{\text{down}}}$, $\bm{T}_{\bm{b}_{\text{down}}}$, $\bm{T}_{\bm{W}_{\text{up}}}$ and  $\bm{T}_{\bm{b}_{\text{up}}}$
are learnable model parameters for the PGN-ADBERT.
For any matrix-format model parameter $\bm{W} \in \mathbb{R}^{M \times N}$,
we have $\bm{T}_{\bm{W}} \in \mathbb{R}^{M \times N \times d}$, where $d$ is the
dim of the annotator embedding.
Similarly, for the vectorial parameter $\bm{b} \in \mathbb{R}^{N}$, we have $\bm{T}_{\bm{b}} \in \mathbb{R}^{N \times d}$.

Thus, the overall input representation of the annotator-adapter can be rewritten as:
\begin{equation}
\label{pgnbert_repr}
\begin{split}
    \bm{r}_1\cdots\bm{r}_n=\text{PGN-ADBERT}(x_1\cdots x_n, \bm{e}^a),
\end{split}
\end{equation}
which jointly encodes the text and the annotator.  % naturally
% The other components of the annotator-adapter model are exactly the same as the
% baseline, including BiLSTM encoding, CRF decoding, and cross-entropy parameter
% learning.

At the training stage, it uses the embedding of crowd annotators to generate
crowd model parameters to learn crowd annotations.
At the inference stage, it uses the centroid point of all annotator embeddings to
estimate the expert, predicting the high-quality opinion expressions for raw texts.
This expert embedding can be computed directly by:
\begin{equation}
    \bm{e}^{\text{expert}} = \frac{1}{|A|} \sum _{a \in A} \bm{e}^a ,
\end{equation}
where $A$ represents all annotators.% who contributed to the training corpus.
% This approximated expert can be interpreted as the elected outcome from
% annotators voting equally.

\subsection{Annotator Mixup}
By scrutinizing the annotator-adapter model,
we can find that there is a minor mismatch during the model training and testing.
During the training, the input annotators are all encoded individually.
While during the testing, the input expert is a mixture of the crowd annotators,
which is never modeled.
To tackle this divergence, we introduce the {\it mixup} \cite{zhang2018mixup}
strategy over the individual annotators to generate a number of synthetic
samples with linear mixtures of annotators, making the training and testing
highly similar.

The mixup strategy is essentially an effective data augmentation method
that has received increasing attention recently in the NLP community
\cite{zhang-etal-2020-seqmix,sun-etal-2020-mixup}.
The method is applied between two individual training instances originally,
by using linear interpolation over a hidden input layer and the output.
In this work, we confine the mixup onto the two training instances with the same
input sentence for annotator mixup.

Formally, given two training instances $(\bm{x}_1 \circ a_1, \bm{y}_1)$ and $(\bm{x}_2 \circ a_2, \bm{y}_2)$,
the mixup is executed only when $\bm{x}_1 = \bm{x}_2$,
thus the interpolation is actually performed between $(a_1, \bm{y}_1)$ and $(a_2, \bm{y}_2)$.
Concretely, the input interpolation is conducted at the embedding layer,
and the output interpolation is directly mixed at the sentence-level:
\begin{equation}
\begin{split}
    & \bm{e}^{\text{mix}} = \lambda \bm{e}^{a_1} + (1 - \lambda) \bm{e}^{a_2}, \\
    & \bm{y}_{\text{mix}} = \lambda \bm{y}_1 + (1 - \lambda) \bm{y}_2,
\end{split}
\end{equation}
where $\lambda \in [0, 1]$ is a hyper-parameter which is usually sampled from the
$\mathrm{Beta}(\alpha, \alpha)$ distribution, and $\bm{y}_{*}$ is the one-hot
vectorial form, where $* \in [1, 2, \text{mix}]$.\footnote{
Note that $\bm{y}_{*}$ is at the sentence-level, where the dimension size is the number of all possible outputs of the given input.
We mix the loss of $\bm{y}_1$ and $\bm{y}_2$ instead of themself in practice.
}
Finally, the loss objective of the new instance is calculated by:
\begin{equation}
\begin{split}
    \mathcal{L}_{\text{mix}} %& = -\log p(Y_{\text{mix}}) \\
     & = -\log \frac{\exp\big(\mathrm{score}(\bm{y}_{\text{mix}})\big)}{\sum_{
        \widetilde{Y}}\exp\big(\mathrm{score}(\tilde{\bm{y}})\big)} ,
\end{split}
\end{equation}
where all scores are computed based on $\bm{x}_1$/$\bm{x}_2$ and $\bm{e}^{\text{mix}}$,
and $\widetilde{Y}$ is all possible outputs for $\bm{x}_1$/$\bm{x}_2$.

Finally, we can produce a number of augmented instances by the annotator mixup.
These instances, together with the original training instances, are used to optimize our model parameters.
The enhanced model is able to perform inference more robustly by using the mixture (i.e, average) of annotators, which is the estimation of the expert.

%%%%%%%%%%%%%%%%%%%%%%%%%%%%%%%%%%%%%%%%%%%%%%%%%%%%%%%%%%%%%%%%%%%%%%%%%%%%%%

% \setlength{\tabcolsep}{6pt}
% \begin{table*}
% \centering \small
% \begin{tabular}{l|ccc|ccc|ccc}
% \toprule
% \multirow{2}{*}{Model} & \multicolumn{3}{c}{Exact}  & \multicolumn{3}{c}{Proportional} & \multicolumn{3}{c}{Binary} \\  % \cline{2-10}
% & P & R & F1 & P & R & F1 & P & R & F1 \\
% \midrule
% MPQA~~~~~~~~~~~~~~~~~~~~~~~~~~~~~
% & 61.06 & 45.49 & 52.14    & 82.47 & 61.44 & 70.42    & 86.98 & 64.80 & 74.27 \\
% \bottomrule
% \end{tabular}
% \caption{The test results of the baseline model (without crowdsourcing adaption)
% on the English MPQA corpus.}
% \label{table:mpqa}
% \end{table*}

\renewcommand{\arraystretch}{0.95}
\setlength{\tabcolsep}{6pt}
\begin{table*}
\centering \small
\begin{tabular}{l|ccc|ccc|ccc}
\toprule
\multirow{2}{*}{Method} & \multicolumn{3}{c|}{Exact}  & \multicolumn{3}{c|}{Proportional} & \multicolumn{3}{c}{Binary} \\  % \cline{2-10}
& P & R & F1 & P & R & F1 & P & R & F1 \\
% \midrule \midrule
% \multicolumn{10}{c}{Chinese OEI (our crowdsourcing dataset)} \\
\midrule
\midrule
Gold    & 61.12 & 53.54 & 57.08    & 81.97 & 72.28 & 76.82    & 85.79 & 77.51 & 81.44 \\
Silver  & 55.27 & 53.25 & 54.24    & 75.79 & 73.01 & 74.37    & 81.23 & 78.25 & 79.71 \\
\midrule
\midrule
ALL     & 61.06 & 45.49 & 52.14    & 82.47 & 61.44 & 70.42    & 86.98 & 64.80 & 74.27 \\
MV      & 53.95 & 50.97 & 52.42    & 74.23 & 70.13 & 72.12    & 78.98 & 74.62 & 76.74 \\
\midrule
LSTM-Crowd \cite{nguyen-etal-2017-aggregating} &
60.55 & 47.68 & 53.35    & \bf 83.79 & 61.32 & 70.82    & \bf 88.71 & 64.92 & 74.98 \\
LSTM-Crowd-cat \cite{nguyen-etal-2017-aggregating} &
59.07 & 47.51 & 52.66    & 77.56 & 62.39 & 69.15    & 83.70 & 67.33 & 74.63 \\
BSC-seq \cite{simpson-gurevych-2019-bayesian} &
40.80 & \bf 59.27 & 48.33 & 55.35 & \bf 82.41 & 66.23 & 60.66 & \bf 90.33 & 72.58 \\
% OptSLA \cite{sabetpour-etal-2020-optsla} &
%  \\
% AggSLC \cite{DBLP:conf/icdm/SabetpourKXL21} &
%  \\
\midrule
Annotator-Adapter \cite{zhang-etal-2021-crowdsourcing} $^\dagger$
& 61.08 & 48.16 & 53.86    & 81.70 & 65.40 & 72.65    & 87.20 & 69.81 & 77.55 \\
Annotator-Adapter + mixup $^\dagger$
& \bf 61.27 & 49.22 & \bf 54.59    & 81.82 & 68.30 & \bf 74.45    & 87.02 & 71.48 & \bf 78.49  \\
% \midrule \midrule
% \multicolumn{10}{c}{English OEI (MPQA 2.0, gold-standard)} \\
% \midrule
% BERT-BiLSTM-CRF (fine-tuning)         & 46.56 & 41.49 & 43.88    & 46.56 & 41.49 & 43.88    & \bf 81.72 & 72.81 & 77.00 \\
% BERT-BiLSTM-CRF (adapter)              & - & - & -    & - & - & -    & - & - & - \\
% % \midrule
% \newcite{xia-etal-2021-unified} + BERT     & \bf 67.15 & 60.63 & 63.71    & - & - & - & - & - & - \\
\bottomrule
\end{tabular}
\caption{The test results, where all methods are backended by BERT-BiLSTM-CRF
for a fair comparison. The \texttt{Gold} and \texttt{Silver} denotes models trained with expert annotations and sentence-level expert aggregation (silver-standard in §\ref{data:silver}), respectively.
The $^\dagger$ indicates statistical significance compared to baselines with $p < 0.01$ by paired t-test.
} \label{table:result}
\end{table*}

\section{Experiment}
\subsection{Setting}
\paragraph{Evaluation.}
We use the span-level precision (P), recall (R) and their F1 for evaluation,
since OEI is essentially a span recognition task.
Following \newcite{breck2007identifying,irsoy-cardie-2014-opinion},
we exploit three types of metrics, namely {\it exact} matching,
{\it proportional} matching and {\it binary} matching, respectively.
The {\it exact} metric is straightforward and has been widely applied for
span-level entity recognition tasks,
which regards a predicted opinion expression as correct only when its start-end
boundaries and polarity are all correct.
Here we exploit the {\it exact} metric as the major method.
The two other metrics are exploited because the exact boundaries
are very difficult to be unified even for experts.
The {\it binary} method treats an expression as correct when it contains an
overlap with the ground-truth expression,
and the {\it proportional} method uses a balanced score by the proportion of
the overlapped area referring to the ground-truth.

We use the best-performing model on the development corpus to evaluate the
performance of the test corpus.
% \footnote{The gold development corpus is only used for the research purpose of fair comparisons, which might be unavailable in practical.}
All experiments are conducted on a single RTX 2080 Ti card at an 8-GPU server with a 14 core CPU and 128GB memory.
We run each setting by 5 times with different
random seeds, and the median evaluation scores are reported.

%where 
%The details are in the Appendix \ref{appendix:evaluation}.

\paragraph{Hyper-parameters.}
We exploit the bert-base-chinese for input representations.\footnote{ %, which consists of 12-layer transformers by the hidden size of 768 for all layers
https://github.com/google-research/bert}
The adapter bottleneck size and the BiLSTM hidden size are set to 128 and 400,
respectively.
For the annotator-adapter, we set the annotator embedding size $d = 8$ and generate the
adapter parameters for the last $p = 6$ BERT layers.
For the annotator mixup, we set $\alpha$ of the $\mathrm{Beta(\alpha,\alpha)}$
distribution to $0.5$.

% We exploit the batched online learning to update the model parameters, with a batch size of 64.
We apply the sequential dropout to the input representations, which randomly
sets the hidden vectors in the sequence to zeros with a probability of $0.2$,
to avoid overfitting.
We use the Adam algorithm to optimize the parameters with a constant learning
rate $1 \times 10^{-3}$ and a batch size $64$, and apply the gradient clipping
mechanism by a maximum value of $5.0$ to avoid gradient explosion.
%and the AdamW method with a learning rate $5 \times 10^{-5}$ and warm-up during
% first 1000 updates for the Adapter layers.

\paragraph{Baselines.}
Two annotator-agnostic baselines (i.e., \texttt{ALL} and \texttt{MV}) and the silver-corpus trained model \texttt{Silver} are all implemented in the same baseline structure and hyper-parameters.
We also implement two annotator-aware methods presented in \newcite{nguyen-etal-2017-aggregating}, where the annotator-dependent noises have been modeled explicitly.
% , namely \texttt{LSTM-Crowd} and \texttt{LSTM-Crowd-cat}.
The \texttt{LSTM-Crowd} model encodes the output
label bias (i.e., noises) for each individual annotator (biased-distributions)
towards the expert (zeroed-distribution), and the \texttt{LSTM-Crowd-cat} model applies
a similar idea but implementing at the BiLSTM hidden layer.
During the testing, zero-vectors are exploited to simulate the expert accordingly.
Their main idea is to reach a robust training on the noisy dataset,
which is totally different from our approach.
In addition, we aggregate crowd labels of the training corpus by a Bayesian inference method \cite{simpson-gurevych-2019-bayesian}, namely \texttt{BSC-seq}, based on their code\footnote{
\href{https://github.com/UKPLab/arxiv2018-bayesian-ensembles}{https://github.com/UKPLab/arxiv2018-bayesian-ensembles}
} and then evaluate its results with the same BERT-BiLSTM-CRF architecture.

\subsection{Main Results}
Table \ref{table:result} shows the test results on our dataset.
In general, the {\it exact} matching scores are all at a relatively low level, demonstrating that precise opinion boundaries are indeed difficult to identify.
With the gradual relaxation of metrics (from {\it exact} to {\it binary}), scores are increased accordingly, showing that these models can roughly locate the opinion expressions to a certain degree.
% The annotator-adapter model can obtain a good performance with an exact matching F1 value of $53.86$, and the relaxed metrics by proportional and binary matching are even much higher.
% When the annotator mixup is applied, the F1 value of the exact matching is boosted (i.e., $54.59-53.86=0.73$), which demonstrates the effectiveness of our proposed annotator mixup.
% The increases in the other two metrics are also similar.
%The potential reason might be due to that the augmented corpus by annotator mixup can let the final model more robust to the expression boundaries.

\paragraph{Dataset comparison.}
Similar to the tasks like NER \cite{zhou-etal-2021-multi}, POS tagging, dependency parsing \cite{straka-2018-udpipe} and so on, in which English models have performed better than the Chinese, we see the same pattern in our OEI task.
The exact matching F1 $57.08$ of the \texttt{Gold} corpus trained model still has a performance gap compared with that of the English MPQA dataset (i.e., $63.71$ by a similar BERT-based model of \newcite{xia-etal-2021-unified}).
This may due to (1) the opinion boundaries in the word-based English MPQA are easier to locate than our character-based Chinese dataset;
(2) the social media domain of our dataset, is more difficult than the news domain of MPQA.
% Notice that due to the Chinese language characteristics, 
% there exist a number of tasks that the performances of their English correspondents (when the training corpus is at a similar scale) 
% are much better such as named entity recognition \cite{zhou-etal-2021-multi}, POS tagging and dependency parsing \cite{straka-2018-udpipe}.

\paragraph{Method comparison.}
First, we compare two annotator-agnostic methods (i.e., \texttt{All} and \texttt{MV})
with annotator-aware ones (i.e., the rest of models).
As shown in Table \ref{table:result}, we can see that annotator-aware modeling is effective as a whole,
bringing better performance on {\it exact} matching.
% The truth-inference method (i.e., \texttt{BSC-seq}) achieve remarkable recalls but its precisions are significantly below the others.
%which seem to be more suitable for our extremely-noisy dataset.
%However, they are merely comparable with the above ones.
% The silver achieves the best results since the expert-selected annotations has the much higher quality.
In particular, our basic annotator-adapter model is able to give the best F1 among these selected baselines, demonstrating its advantage in crowdsourcing modeling.
When the annotator-mixup is applied, the test scores are further boosted,
% we can obtain an improved F1 value by the {\it exact} matching (i.e., $54.59-53.86=0.73$),
showing the effectiveness of our annotator mixup.
% , where the final number is comparable to the \texttt{Silver} model.
% The observation shows the effectiveness of our annotator mixup.
The overall tendencies of the two other metrics are similar by comparing our models with the others.

Our final performance is not only comparable to the \texttt{silver} corpus trained model, which we can take it as a weak upper-bound.
but also close to the upper-bound model with expert annotations (i.e., \texttt{Gold}).
Thus, our result for Chinese OEI is completely acceptable, demonstrating that crowdsourcing annotations are indeed with great value for model training.
The observation indicates that crowdsourcing could be a highly-promising alternative to build a Chinese OEI system at a low cost.
%The characteristic is very important especially for a new-coming domain in Chinese OEI.
%All these

\subsection{Analysis}
Here we conduct fine-grained analyses to better understand the task and these methods in-depth, where the evaluation by {\it exact} matching is used in this subsection.
There are several additional analyses which are shown in the Appendix.

\begin{figure}\centering
\includegraphics[scale=1.05]{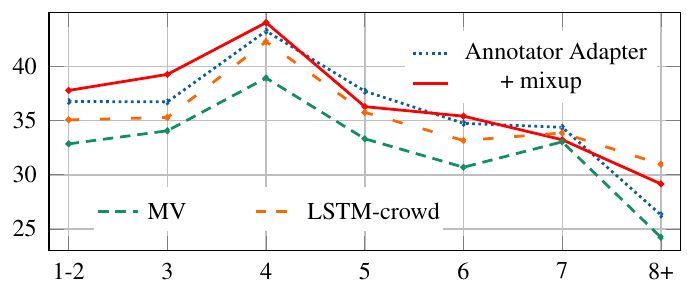}
\caption{F1 scores of {\it exact} matching in terms of the opinion expression length.
We bucket the opinion expressions into seven categories, where each category
includes more than 100 opinion expressions.}
\label{figure:span-recall}
\end{figure}

\paragraph{Performance by the opinion expression length.}
Intuitively, the identification of opinion expressions can be greatly affected
by the length of the expressions, and longer expressions might be more challenging
to be identified precisely.
Figure \ref{figure:span-recall} shows the F1 scores in terms of expression lengths
by the four models we focused.
We can see that the F1 score decreases dramatically when the expression
length becomes larger than 4, which is consistent with our intuition.
In addition, the annotator-adapter model is better than previous methods,
and the mixup model can reach the best performance on almost all the categories,
indicating the robustness of our annotator mixup.
%And the spans of length 4 have the highest F1, it is because this kind of spans
% has the largest proportion in the dataset and are mostly high-quality and easy
% to identify by our observation.

\begin{figure}\centering
\includegraphics[scale=0.65]{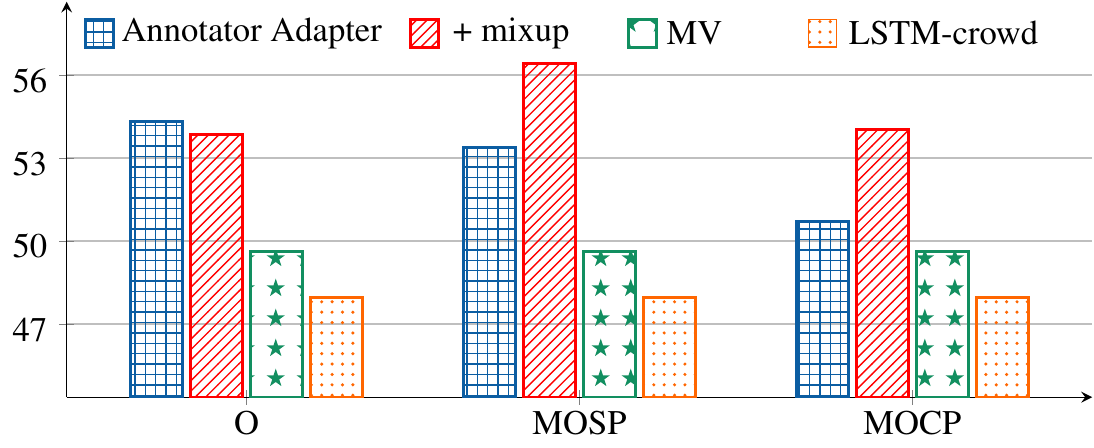}
\caption{F1 scores of {\it exact} matching by following three category sentences:
(1) one-opinion ({\bf O}),
(2) multiple-opinion single-polarity ({\bf MOSP}),
and (3) multiple-opinion contradict-polarity ({\bf MOCP}).
}
\label{figure:kind-score}
\end{figure}

\paragraph{Influence of the opinion number per sentence.}
One sentence may have more than one opinion expressions,
where these opinions might be mutually helpful or bring increased ambiguities.
It is interesting to study the model behaviors in terms of opinion numbers.
Here we conduct experimental comparisons by dividing the test corpus into three categories:
(1) only one opinion expression exists in a sentence;
(2) at least two opinions exist, and they are of the same sentiment polarity;
(3) both positive and negative opinion expressions exist.
As shown in Figure \ref{figure:kind-score}, the sentences with multiple opinions
of a consistent polarity can obtain the highest F1 score.
The potential reason might be that the expressed opinions of these sentences
are usually highly affirmative with strong sentiments,
and the consistent expressions can be mutually helpful according to our assumption.
For the other two categories,
it seems that they are equally difficult according to the final scores.
For all three categories, two annotator-adapter models demonstrate better
performance than the others.

%Then the texts with one opinion slightly lower than the former.
%Although they are the largest part (855 of 1517), but their emotions are not so intense as the former class.
%The sentiment flip in one text is the most difficult situation and have fewest instances, got the worst performance.

%3、转折，句子中有不同情感，看这些的准确率，一个、多个一致、多个不一致。（多个低于一个的，存在
%情感不一致，难度高）。测试集opinion数-句子数 1: 855, 2: 504, 3: 129, 4: 23, 5: 6,
%单句中极性种类1: 1391, 2: 126。柱状图就可

%简单分析每个人的性能。一个人好不只是标得好，也是从其他人学的好。
%1、测试集。假设用户给定，测准确率，同一个用户，随机交叉用户。

\setlength{\tabcolsep}{6pt}
\begin{table}
\centering \small
\begin{tabular}{l|ccc}
\toprule
\multirow{2}{*}{Model} & \multicolumn{3}{c}{Exact} \\  % \cline{2-10}
  & P & R & F1 \\
\midrule
ALL                 & 52.24 & 34.17 & 41.32 \\
MV                  & 43.79 & 38.70 & 41.09 \\
% \midrule
% Silver              &  &  &  \\
\midrule
LSTM-Crowd          & 46.57 & 38.19 & 41.97 \\
LSTM-Crowd-cat      & 52.10 & 32.79 & 40.25 \\
\midrule
Annotator-Adapter          & \bf 55.81 & 42.80 & \bf 48.45 \\
Annotator-Adapter + mixup  & 52.76 & \bf 43.68 & 47.79 \\
\bottomrule
\end{tabular}
\caption{The evaluation results on the crowd test set, i.e., we compute F1
scores between model predictions and crowd annotations. The ALL and MV have no
modifications. The other annotator-aware models have replaced the expert vector
with the specific annotator vector corresponding to the annotations when testing.}
\label{table:self}
\end{table}

\paragraph{Self-evaluation of crowd annotators.}
The annotator adapter uses a pseudo expert embedding to predict
opinion expressions and evaluate performance on the gold-standard annotations of experts.
It is interesting to examine the self-evaluation performance on
the crowd annotations of the test corpus as well.
During the inference, we use the crowd annotators as inputs, and calculate the
model performance on the corresponding crowd annotations.

Table \ref{table:self} shows the results.
First, two annotator-agnostic models (i.e., \texttt{ALL} and \texttt{MV}) have similar poor performance since they are trying to estimate the expert annotation function rather than learn crowd annotations.
Second, the performance of two annotator-noise-modeling methods, \texttt{LSTM-Crowd} and \texttt{LSTM-Crowd-cat}, respectively, is close to the annotator-agnostic ones, showing that they are also incapable to model individual annotators.
Then, our two annotator-adapter models achieve leading performance compared with all baseline methods, giving a significant gap (at least $47.79-41.97=5.82$ in F1).
They are more capable of predicting crowd annotations, demonstrating the ability to model the annotators effectively.
To our surprise, the mixup annotator-adapter model does not exceed the basic one, indicating that the mixed annotator embeddings in training could slightly hurt the modeling of individual annotators.

\section{Related Work}
% \subsection{Opinion Expression Identification}
% Opinion mining has been one top topic in the NLP community, which involves a wide of range of tasks \cite{liu2012sentiment}.
OEI is one important task in opinion mining \cite{liu2012sentiment},
and has received great interests \cite{breck2007identifying,irsoy-cardie-2014-opinion,xia-etal-2021-unified}.
The early studies can be dated back to \newcite{wilson-etal-2005-opinionfinder}
and \newcite{breck2007identifying},
which exploit CRF-based methods for the task with manually-crafted features.
SemiCRF is exploited next in order to exploit span-based features \cite{yang-cardie-2012-extracting}.
Recently, neural network models have attracted the most attention.
\newcite{irsoy-cardie-2014-opinion} present a deep bi-directional recurrent
neural network (RNN) to identify opinion expressions.
% \newcite{wang-etal-2016-recursive} use dependency tree based RNN to learn the
% feature representations for Aspects and Opinions Co-Extraction.
BiLSTM is also used in \newcite{katiyar-cardie-2016-investigating} and \newcite{ZHANG201956},
showing improved performance on OEI.
\newcite{fan-etal-2019-target} design an Inward-LSTM to incorporate the opinion
target information for identifying opinion expressions given their target,
which can be seen as a special case of our task.
% \newcite{wang-pan-2018-recursive} study the unsupervised domain adaptation for aspect
% and opinion terms extraction, and \newcite{zhao-etal-2020-spanmlt} proposed a
% span-based multi-task model for the same task.
\newcite{xia-etal-2021-unified} employ pre-trained BERT representations
\cite{devlin-etal-2019-bert} to increase the identification performance of joint extraction of the opinion expression, holder and target by a span-based model.

All the above studies are in English and based on the MPQA \cite{wiebe2005annotating},
or customer reviews \cite{wang-etal-2016-recursive,wang2017coupled,fan-etal-2019-target}
since there are very few datasets available for other languages.
Hence, we construct a large-scale Chinese corpus for this task by crowdsourcing,
and borrow a novel representation learning model \cite{zhang-etal-2021-crowdsourcing} to handle the crowdsourcing annotations.
In this work, we take the general BERT-BiLSTM-CRF architecture as the baseline, which is a competitive model for OEI task.
% We follow \newcite{katiyar-cardie-2016-investigating} using BiLSTM-CRF as the
% baseline architecture and enhance it with pre-trained BERT
% \cite{devlin-etal-2019-bert}, which can result in the state-of-the-art performance
% in the literature.

% \subsection{Crowdsourcing}
Crowdsourcing as a cheap way to collect a large-scale training corpus for
supervised models has been gradually popular in practice
\cite{snow-etal-2008-cheap,callison-burch-dredze-2010-creating,dietrich2020fine}.
A number of models are developed to aggregate a higher-quality corpus from the
crowdsourcing corpus \cite{Raykar-2010-Learning,rodrigues2014gaussian,rodrigues2014sequence,Moreno-2015-Bayesian},
aiming to reduce the gap over the expert-annotated corpus.
Recently, modeling the bias between the crowd annotators and the oracle experts
has been demonstrated effectively
\cite{nguyen-etal-2017-aggregating,simpson-gurevych-2019-bayesian,li-etal-2020-neural},
focusing on the label bias between the crowdsourcing annotations and gold-standard answers,
regarding crowdsourcing annotations as annotator-sensitive noises.
\newcite{zhang-etal-2021-crowdsourcing} do not hold crowdsourcing annotations as noisy
labels, while regard them as ground-truths by the understanding of individual crowd annotators.
In this work, we follow the idea of \newcite{zhang-etal-2021-crowdsourcing} to explorate our crowdsourcing corpus, and further propose the annotator mixup to enhance the learning of the expert representation for the test stage.
%\newcite{nguyen-etal-2017-aggregating} has mentioned an idea most close to ours,
%but the presented model (referred to as LSTM-Crowd-cat) is very initial,
%which is served as a negative baseline in their work.

% conclusion两段
\section{Conclusion}
We presented the first work of Chinese OEI by crowdsourcing, which is also the first crowdsourcing work of OEI.
First, we constructed an extremely-noisy crowdsourcing corpus at a very low cost, and also built gold-standard dataset by experts for experimental evaluations.
To verify the value of our low-cost and extremely-noisy corpus, we exploited the annotator-adapter model presented by \newcite{zhang-etal-2021-crowdsourcing} to fully explore the crowdsourcing annotations,
and further proposed an annotator-mixup strategy to enhance the model.
Experimental results show that the annotator-adapter can make the best use of our crowdsourcing corpus compared with several representative baselines, and the annotator-mixup strategy is also effective. Our final performance can reach an F-score of $54.59\%$ by exact matching.
This number is actually highly competitive by referring to the model trained on expert annotations ($57.08\%$),
which indicates that crowdsourcing can be highly recommendable to set up a Chinese OEI system fast and cheap, although the collected corpus is extremely noisy.
\section*{Ethical/Broader Impact}
We construct a large-scale Chinese opinion expression identification dataset with crowd annotations.
We access the original posts by manually traversing the relevant Weibo topics or searching the corresponding keywords, and then copy and anonymize the text contents.
All posts we collected are open-access.
In addition, we also anonymize all annotators and experts (only keep the ID for the research purpose).
All annotators were properly paid by their actual efforts.
This dataset can be used for both the Chinese opinion expression identification task as well as crowdsourcing sequence labeling.

\section*{Acknowledgments}
We thank all reviewers for their hard work.
This research is supported by grants from
the National Key Research and Development Program of China (No. 2018YFC0832101)
and the National Natural Science Foundation of China (No. 62176180).

\bibliography{ref}
\bibliographystyle{acl_natbib}

\newpage
\appendix

% 把gudeline例子去掉
\section{Annotation Guideline}\label{appendix:guideline}
In this annotation task, we will give a number of sentences that have a high probability of expressing positive or negative sentiment, and your goal is to label the words that expresses these sentiments in each sentence.
An intuitive criteria for determining whether words are expressing sentiment is that if these words are replaced, the sentiment expressed by the sentence will also change.
Sentimental words will not usually be names of people, places, time or pronouns, etc.
It is important to note that (1) you need to carefully understand the emotion expressed by the sentence, not judge it according to your own values, and
(2) the labeled words usually do not include the target of the sentiment, such as pronouns, names of people, etc., which are generally not affected by the replacement of these words.
% Below we will give some specific examples. (the {\bf bold} words express positive sentiment, {\it italic} ones express negative sentiment.)

% \zx{搞一些容易混淆的}
% \begin{enumerate}
% \item I am {\it not destined to be happy} for the rest of my life.
% % \item Dr. Li Wenliang's last words, {\it tears}!
% \item Under the COVID-19, there is a man who is deservedly the ``{\bf Idol of all people}'', his name is Zhong Nanshan.
% \item Ms. Wang Fang is {\it just too smart}.
% \item Fang Fang {\it won}. The efforts of 1.4 billion Chinese can't compare to this little diary of a closed city.
% % \item When you reach middle age, your dreams are almost always scenes of your past work, life and study, which seem to be {\it more than reality...}
% \item The government has done something {\bf very impressive}: the contact
% information of more than 2,000 community secretaries in the city has been made public.
% \item I don’t know {\it how much perseverance} it takes to revisit the past on Weibo and WeChat one by one.
% \item I {\it really have to admire} Fang Fang, this way of writing perfectly avoids the risk.

% \item ......
% \end{enumerate}

% \section{Evaluation}\label{appendix:evaluation}
% \subsection{Evaluation}

\begin{table}
\centering \small
\begin{tabular}{l|ccc}
\toprule
Model & Exact F1 & Prop F1 & Binary F1 \\
\hline
\hline
\multicolumn{4}{c}{$p$} \\ \hline
    4  & 46.00 & 60.95 & 65.02 \\
    \bf 6  &  \bf 53.86 & \bf 72.65 & 77.55 \\
    8  & 53.39 & 72.54 & \bf 78.32 \\
    10 & 53.21 & 72.55 & 78.08 \\
    12 & 52.82 & 71.04 & 74.52 \\
    \hline
\hline
\multicolumn{4}{c}{$\alpha$ in $\mathrm{Beta}(\alpha, \alpha)$} \\ \hline
0.2 & 54.86 & 72.92 & 78.19 \\
\bf 0.5 & 55.15 & \bf 73.37 & \bf 77.79 \\
0.8 & \bf 55.18 & 73.00 & 77.48 \\
1.0 & 54.78 & 72.35 & 76.98 \\
\hline
\hline
\multicolumn{4}{c}{Mixup Training Strategy} \\ \hline
One-Stage &  \bf 55.15 & 73.37 & 77.79 \\
\bf Two-Stage &  54.59 & \bf 74.45 & \bf 78.49 \\
\hline
\hline
\multicolumn{4}{c}{Fine-tuning Based Models} \\ \hline
ALL     & 52.35 & 69.58 & 76.99    \\
MV      & 47.52 & 69.07 & 76.09    \\
Silver  & \bf 54.47 & \bf 73.16 & \bf 79.99    \\
(2017)  & 53.17 & 70.81 & 77.23    \\
(2017)-cat  & 53.01 & 70.03 & 76.95    \\
% \hline
% \hline
% \multicolumn{4}{c}{BERT-BiLSTM-CRF on MPQA 2.0} \\ \hline
% Fine-tuning     &   &   &   \\
% Adapter-tuning     &   &   &   \\
\bottomrule
\end{tabular}
\caption{Experimental results of various settings.}
\label{table:tuning}
\end{table}

\section{Hyper-parameter Tuning}
% PGN layer 调参，network size
% mixup的pre train 和 直接 train（各ann性能都强，估计都比较像了）
%\paragraph{Network Size}
%First, we explore different adapter bottleneck and LSTM hidden size settings,
%found that the LSTM size has no significant impact on performance, but the
%adapter size must match the dataset size to achieve better performance.
% 5、pgn-adapter层数。
We also implement the baseline models in the fine-tuning style, results (in
Table \ref{table:tuning}) show that the adapter-based models are comparable and
parameter-efficient.

\paragraph{PGN Adapter Layers}
First, we examine the influence of PGN adapter layers mentioned in §\ref{sec:annotator-adapter}
by $p$, which is a hyper-parameter in our annotator-adapter.
% We traverse the value by 4, 6, 8, 10, and 12,
% and the results are shown in Table \ref{table:tuning}.
As shown in Table \ref{table:tuning}, we can see that the performance is stable between $p \in [6, 8, 10]$.
After considering both the parameter scale and the capability of our model,
we set $p=6$ for a trade-off.

% more layers we applied, better performance we gain.
% 4没效果，6 8 可以，10 12 又低一些
%And more layers we applied,
%For the trade-off between training parameters and model ability, we choose to
%insert adapter in half of BERT layers in the main experiments.

% \paragraph{Parameter Sharing}
% Additionally, we tried to share adapter parameters across transformer layers,
% which can significantly reduce parameters.
% Results shows a performance degrade about 3-5 F1 score.

\paragraph{Annotator Mixup}
The mixup includes a hyper-parameter $\alpha$ to control the interpolation by
the distribution $\mathrm{Beta}(\alpha, \alpha)$.
Here we show the influence of $\alpha$ by setting it with 0.2, 0.5, 0.8, and 1.0.
We find that the model performance has no significant differences between these values,
as shown in Table \ref{table:tuning}.
To train our mixup model,
we also have a reasonable small trick: training the mixup model in two stages.
First, the model is trained only with the original corpus.
When the model achieves the best performance on the devset,
we begin the second-stage training by using the original corpus as well as the augmented corpus.
Their performance difference is shown in Table \ref{table:tuning},
which indicates that the two-stage training is important for our mixup model.
%In our practice, we initiate the weights from the checkpoint of vanilla model
%at epoch 1-5 (in most cases, 2) depends on development result.
%For the $\mathrm{Beta}$ distribution parameter $\alpha$, we tried multiple
%values ​​from 0.2 to 1, as Table \ref{table:tuning} shows, it achieved the best
%result at 0.5, which we used in our final result.
%Although mixup training from scratch is feasible, its performance is lower than
%the above continue training, but slightly higher than the vanilla model (as
%shown in Table \ref{table:tuning}).
%The mixup training is quite a simple and convenient strategy, we believe that
%combined with more sophisticated tuning methods, the model can improve performance further.

% 解释一下为什么不稳定
\section{Expert-Evaluation of Crowd Annotators}
We evaluate the performance of each learned annotator of three annotator-aware models towards the expert's view.
The goal is achieved by using the individual annotator embeddings as input to 
obtain the output predicted by this specific annotator,
and then measure the output performance based on the gold-standard test corpus.
\tabref{table:annotator} shows the results.
There is a huge discrepancy between the scores of different learned annotators of \texttt{LSTM-Crowd} or \texttt{annotator-adapter}, demonstrating annotators have different abilities in predicting gold labels.
This is mainly because the annotators have different abilities meanwhile the annotations they gave have different qualities.
All annotators in the annotator-adapter model are unable to outperform the expert (centroid point), verifying that the estimated expert is strong and reasonable.
In addition, the learned annotators of our mixup model have closer performances since the annotator-mixup change the learning objective from modeling annotators to modeling the expert, which can further boost the performance of the estimated expert.
% 这里实际上有好多 annotator 超过了 mixup 的 expert, 可以当做 future work

\begin{table}[t] % [htpb]
\centering
\includegraphics[scale=0.73]{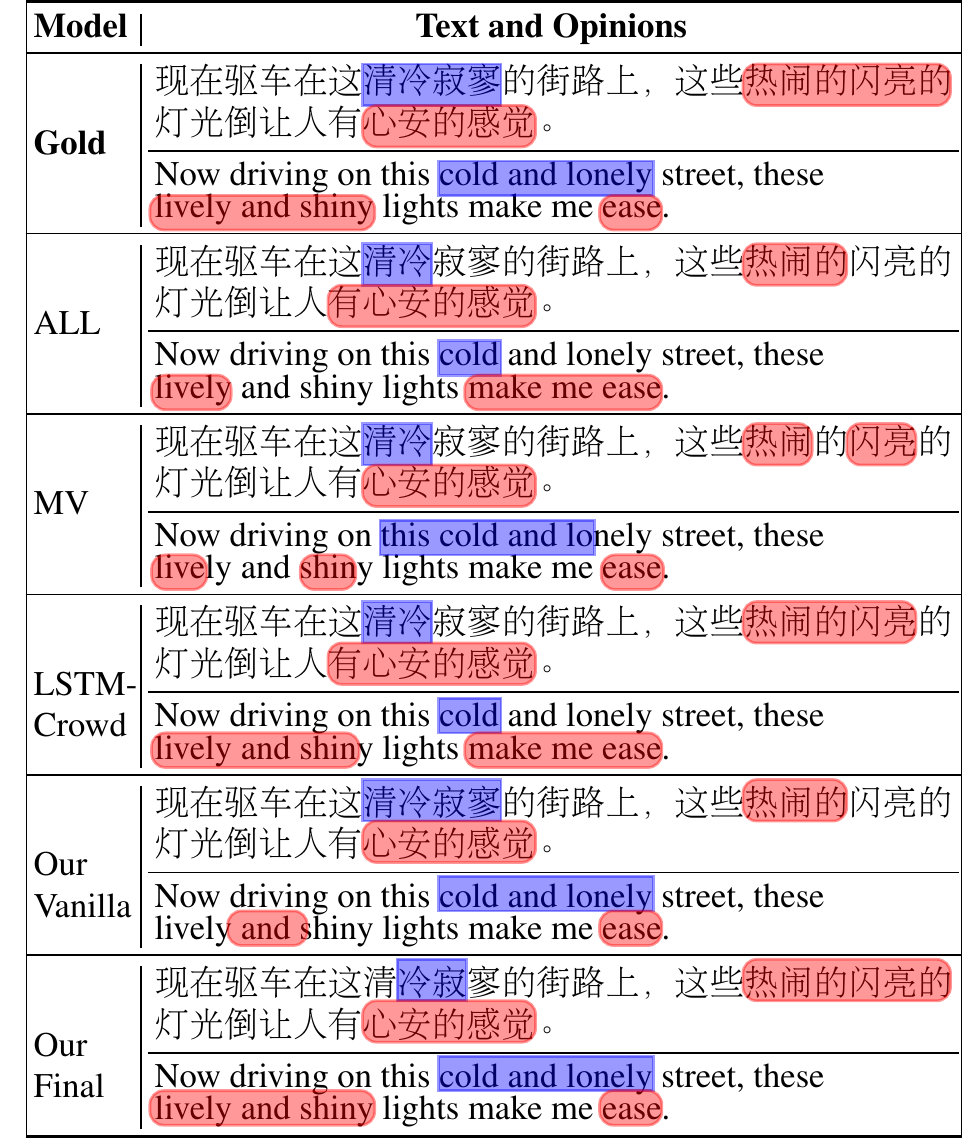}
\caption{
Case Study. The blue rectangles and red boxes with round corners are negative and
positive, respectively.
% \texttt{AA} represents Annotator-Adapter.
}
\label{table:case}
\end{table}

\begin{table*} \centering
\resizebox*{\textwidth}{!}{
\begin{tabular}{cccc|cccc|cccc}
\toprule
Annota- & LSTM- & Annotator- & + &
Annota- & LSTM- & Annotator- & + &
Annota- & LSTM- & Annotator- & + \\
tor ID& Crowd & Adapter & mixup &
tor ID& Crowd & Adapter & mixup &
tor ID& Crowd & Adapter & mixup \\
\midrule
0  & 50.76 & 47.31 & 54.63 &  24 & 28.45 & 11.99 & 44.04 &  48 & 40.27 & 40.52 & 55.27 \\
1  & 44.02 & 40.05 & 51.84 &  25 & 51.69 & 47.50 & 55.56 &  49 & 47.28 & 50.45 & 54.80 \\
2  & 53.30 & 48.20 & 55.18 &  26 & 49.05 & 40.18 & 53.26 &  50 & 51.27 & 47.89 & 52.72 \\
3  & 38.63 & 13.01 & 45.14 &  27 & 51.58 & 48.08 & 54.19 &  51 & 50.86 & 49.45 & 54.42 \\
4  & 43.37 & 29.78 & 55.22 &  28 & 51.69 & 38.10 & 48.99 &  52 & 54.92 & 40.82 & 49.88 \\
5  & 55.02 & 47.95 & 53.84 &  29 & 51.46 & 46.31 & 55.06 &  53 & 47.63 & 31.20 & 52.81 \\
6  & 45.02 & 46.13 & 54.66 &  30 & 45.30 & 33.83 & 55.20 &  54 & 49.60 & 43.54 & 54.85 \\
7  & 52.93 & 43.56 & 55.60 &  31 & 46.19 & 44.14 & 49.29 &  55 & 54.88 & 41.97 & 55.44 \\
8  & 35.40 & 22.55 & 46.86 &  32 & 50.02 & 41.63 & 53.52 &  56 & 55.98 & 52.35 & 56.12 \\
9  & 46.61 & 37.30 & 54.58 &  33 & 36.78 & 40.17 & 54.43 &  57 & 53.56 & 44.90 & 53.13 \\
10 & 50.33 & 45.37 & 54.76 &  34 & 39.01 & 34.48 & 52.80 &  58 & 45.19 & 31.42 & 48.81 \\
11 & 49.98 & 48.87 & 54.17 &  35 & 48.17 & 49.09 & 52.66 &  59 & 53.09 & 43.95 & 53.65 \\
12 & 53.90 & 48.53 & 55.69 &  36 & 54.45 & 47.14 & 56.18 &  60 & 35.27 & 13.13 & 52.97 \\
13 & 54.51 & 49.11 & 54.88 &  37 & 53.32 & 43.87 & 54.44 &  61 & 52.46 & 34.26 & 54.79 \\
14 & 49.86 & 48.65 & 53.08 &  38 & 51.08 & 43.25 & 52.08 &  62 & 41.95 & 38.49 & 51.39 \\
15 & 41.64 & 32.81 & 49.25 &  39 & 42.33 & 31.08 & 52.68 &  63 & 35.73 & 43.76 & 54.10 \\
16 & 53.51 & 41.33 & 53.95 &  40 & 46.63 & 42.81 & 53.46 &  64 & 52.56 & 40.93 & 52.93 \\
17 & 50.11 & 34.24 & 52.71 &  41 & 46.50 & 40.38 & 53.45 &  65 & 48.70 & 34.95 & 51.83 \\
18 & 52.80 & 41.83 & 54.98 &  42 & 50.31 & 44.68 & 51.85 &  66 & 46.21 & 30.29 & 52.67 \\
19 & 42.29 & 35.71 & 51.46 &  43 & 54.73 & 48.57 & 51.47 &  67 & 46.24 & 33.75 & 49.53 \\
20 & 51.38 & 47.30 & 52.00 &  44 & 47.34 & 31.86 & 52.75 &  68 & 35.36 & 15.34 & 50.08 \\
21 & 35.39 & 37.10 & 47.02 &  45 & 46.83 & 28.98 & 54.59 &  69 & 32.06 & 22.67 & 52.54 \\
22 & 52.62 & 43.67 & 53.10 &  46 & 54.26 & 40.30 & 52.05 &
\multirow{2}{*}{Expert} & \multirow{2}{*}{53.35} & \multirow{2}{*}{53.86} & \multirow{2}{*}{54.59} \\
23 & 53.49 & 47.08 & 54.62 &  47 & 49.73 & 41.55 & 54.41 & \\
\bottomrule
\end{tabular}
}
\caption{
The F1 scores by using different crowd annotators as input on the gold testset. Exact matching scores are reported. The LSTM-Crowd just learns an estimation of expert assisted by modeling the label bias of annotators, while the annotator-adapter model learns the different understandings of each annotator but not the expert annotations.
% The LSTM-Crowd and annotator-adapter models have their respective advantages,
% and 
Our final mixup model is much more stable across different annotators.
The observation indicates that, with the application of annotator-mixup, all annotators can learn from each other and improve towards the expert level together, which can enhance the expert-modeling.
% Combining the relativly lower performance in the annotator-self-evaluation of our mixup model,
% thus the annotator-modeling objective is transformed to the expert-modeling.
}\label{table:annotator}
\end{table*}

\section{Case Study}
For a more intuitive understanding of our task and various models,
we offer a paradigmatic example from the test set to analyze their outputs.
Table \ref{table:case} shows the gold annotation and model predictions.
As shown, the \texttt{ALL} method can correctly recognize all three opinions,
but fails to predict the correct boundaries.
The \texttt{MV} method splits one opinion into two, and is able to recall one full opinion expression exactly.
The \texttt{LSTM-Crowd} is similar to \texttt{ALL} yet slightly better.
Both the annotator-adapter and our mixup models can obtain better results for this example.
Note that all three opinions are difficult to be fully recognized even by crowd annotators.

\end{CJK}
\end{document}